%% file: main.tex
\documentclass[sigconf]{acmart}
\AtBeginDocument{%
  }

\setcopyright{acmlicensed}
\copyrightyear{2018}
\acmYear{2018}
\acmDOI{XXXXXXX.XXXXXXX}
\acmConference[Conference acronym 'XX]{Make sure to enter the correct
  conference title from your rights confirmation email}{June 03--05,
  2018}{Woodstock, NY}
\acmISBN{978-1-4503-XXXX-X/2018/06}

\usepackage{multirow}
\usepackage{enumitem}
\usepackage{stfloats}
\begin{document}

\title{TransLLM: A Unified Multi-Task Foundation Framework for Urban Transportation via Learnable Prompting}


\author{Jiaming Leng}
\affiliation{%
  \institution{University of Science and Technology of China}
  \country{}}
\email{lengjm@mail.ustc.edu.cn}

\author{Yunying Bi}
\affiliation{%
  \institution{University of Science and Technology of China}
  \country{}}
\email{biyunying@mail.ustc.edu.cn}

\author{Chuan Qin}
\affiliation{%
 \institution{Computer Network Information Center, Chinese Academy of Sciences.}
 \country{}}
 \email{chuanqin0426@gmail.com}

\author{Bing Yin*}
\affiliation{%
  \institution{iFLYTEK}
  \country{}}
  \email{bingyin@iflytek.com}

\author{Yanyong Zhang}
\affiliation{%
  \institution{University of Science and Technology of China}
  \country{}}
\email{yanyongz@ustc.edu.cn}

\author{Chao Wang*}
\affiliation{%
  \institution{University of Science and Technology of China}
  \country{}}
\email{wangchaoai@ustc.edu.cn}


\input{Chapters/Abstract}







\newcommand{\doubleunderline}[1]{\underline{\underline{#1}}}

\maketitle
\input{Chapters/Introduction}

\input{Chapters/Relatedwork}
\input{Chapters/Preliminaries}

\input{Chapters/Method}

\input{Chapters/Experiment}

\input{Chapters/Conclusion}

\bibliographystyle{ACM-Reference-Format}
\bibliography{sample-base}

\input{Chapters/Appendix}

\end{document}

%% file: Chapters/Abstract.tex
\begin{abstract}
  Urban transportation systems encounter diverse challenges across multiple tasks, such as traffic forecasting, electric vehicle (EV) charging demand prediction, and taxi dispatch. Existing approaches suffer from two key limitations: small-scale deep learning models are task-specific and data-hungry, limiting their generalizability across diverse scenarios, while large language models (LLMs), despite offering flexibility through natural language interfaces, struggle with structured spatiotemporal data and numerical reasoning in transportation domains. To address these limitations, we propose TransLLM, a unified foundation framework that integrates spatiotemporal modeling with large language models through learnable prompt composition. Our approach features a lightweight spatiotemporal encoder that captures complex dependencies via dilated temporal convolutions and dual-adjacency graph attention networks, seamlessly interfacing with LLMs through structured embeddings. A novel instance-level prompt routing mechanism, trained via reinforcement learning, dynamically personalizes prompts based on input characteristics, moving beyond fixed task-specific templates. The framework operates by encoding spatiotemporal patterns into contextual representations, dynamically composing personalized prompts to guide LLM reasoning, and projecting the resulting representations through specialized output layers to generate task-specific predictions. Experiments across seven datasets and three tasks demonstrate the exceptional effectiveness of TransLLM in both supervised and zero-shot settings. Compared to ten baseline models, it delivers competitive performance on both regression and planning problems, showing strong generalization and cross-task adaptability. Our code is available at https://github.com/BiYunying/TransLLM.
  \vspace{-9.5pt}
\end{abstract}

%% file: Chapters/Introduction.tex
\vspace{-1em}
\section{Introduction}
Urban transportation systems are the lifelines of modern cities, yet they face persistent challenges arising from increasing travel demand, dynamic spatiotemporal patterns, and the pressing need for efficient resource management. In this context, common tasks include forecasting traffic flow to alleviate congestion~\cite{li2017diffusion,wu2019graph}, predicting charging demand to support the development of electric vehicle infrastructure~\cite{yi2022electric,wang2023short}, and optimizing taxi dispatch to balance supply and demand across urban regions~\cite{yao2018deep,liu2020context}. While these tasks have distinct objectives, they are inherently interconnected through shared spatiotemporal dynamics and urban mobility patterns~\cite{zhang2017deep}, requiring a unified modeling framework that can capture cross-task dependencies and leverage shared knowledge across different transportation domains. 

Traditional approaches address each task in isolation using handcrafted rules or rigid statistical models~\cite{5940418,KUMAR2017582,SMITH2002303}, failing to capture the complex nonlinear dynamics and cross-task dependencies in modern transportation systems. Driven by advances in deep learning, a multitude of specialized small-scale models have emerged for urban transportation applications. Notable among these are GNN-based models ~\cite{guo2019attention,fan2025pdg2seq,fang2021spatial}, which leverage meticulously designed modules to effectively capture complex spatial dependencies. However, these models have significant limitations. First, they heavily rely on extensive labeled data, which can be costly and impractical to collect across diverse urban environments, especially when transferring to new cities without existing annotated datasets. Second, their task-specific designs inherently restrict generalization, making it challenging to adapt them to a wide array of transportation problems. In addition, their rigid architectures limit the integration of heterogeneous external information ~\cite{yao2019revisiting} such as sampling frequency and fine-grained temporal details, both of which are crucial for building robust and flexible transportation forecasting systems.

The emergence of large models has brought new opportunities for modeling urban transportation systems. Large language models such as GPT-4~\cite{achiam2023gpt}, LLaMA~\cite{grattafiori2024llama}, and Gemini~\cite{team2023gemini} can easily integrate external information, including fine-grained temporal details, the sampling frequency of historical data, and urban points of interest (POIs), using their natural language understanding capabilities. By utilizing natural language interfaces, these models offer greater scalability and interactivity. However, current LLMs often lack adequate numerical reasoning capabilities in the spatiotemporal domain~\cite{NEURIPS2023_deb3c281}, leading to prediction biases or outputs that violate basic physical constraints. For example, continuous numerical inputs such as traffic flow or travel time must be tokenized into discrete text, which can introduce precision loss and hinder the model's ability to capture fine-grained numerical patterns. Recent studies have explored hybrid approaches that combine large and small models, demonstrating promising potential across various application domains such as recommendation systems~\cite{10.1145/3604915.3610647,wang2024nextgenerationllmbasedrecommendersystems}, intelligent healthcare~\cite{10.1145/3661167.3661202,gao2024guidingiotbasedhealthcarealert}, cognitive robotics~\cite{10802322,ly2025inteliplaninteractivelightweightllmbased}. In the transportation domain, UrbanGPT~\cite{li2024urbangpt} leverages a multi-level temporal convolutional network to enhance the LLM’s understanding of sequential patterns, while LLMLight~\cite{lai2023llmlight} constructs task-specific prompt templates to guide LLM reasoning and improve interpretability. However, these methods are limited to simple regression or classification tasks and struggle to handle more complex planning problems, such as taxi dispatching. Moreover, they rely on fixed, task-wise prompt templates, which constrain the model’s ability to adapt to diverse and heterogeneous data scenarios~\cite{mao2025reinforced}.

Despite recent progress, several key challenges remain in building a general-purpose traffic foundation model. First, multi-task transportation scenarios involve heterogeneous spatiotemporal inputs~\cite{9204396}—such as traffic flow, travel demand, and geographic adjacency—which are often structured and difficult to represent in natural language. Second, even within a single task, input instances may exhibit diverse temporal and spatial dynamics.Existing task-wise prompting strategies lack adaptability to intra-task variation, making it difficult to handle sample-level differences caused by regions, time periods, or contextual changes. Third, a unified traffic model must contend with the diversity of task objectives, output formats, and potential interference among tasks, posing significant challenges to model generalization and stability~\cite{ruder2017overview}.

To enable the understanding of structured data, we design a general spatiotemporal dependency encoder that transforms structured traffic signals into embeddings, which can be seamlessly integrated into prompts. To handle instance-level variability in input samples, we introduce learnable prompts that adaptively select the most suitable prompt for each instance. To support multiple tasks, we avoid letting the LLM generate outputs directly; instead, we employ a multi-task output layer tailored to each prediction target. In summary, TransLLM operates by encoding spatiotemporal patterns into contextual representations, dynamically composing personalized prompts to guide LLM reasoning, and projecting the resulting representations through specialized output layers to  generate task-specific predictions.
Our key contributions are:
\vspace{-0.5em}
\begin{itemize}[leftmargin=*]
\item We propose TransLLM, a unified foundation framework that integrates spatiotemporal encoding with large language models for diverse transportation tasks, supporting both forecasting and optimization problems.

\item We design a lightweight spatiotemporal encoder combining dilated temporal convolutions with dual-adjacency graph attention networks, enabling task-agnostic modeling of complex spatiotemporal dependencies.

\item We introduce a learnable instance-level prompt routing mechanism using reinforcement learning, which dynamically assembles personalized prompts based on input characteristics, moving beyond fixed task-wise templates.

\item Comprehensive experiments on seven datasets across three transportation tasks demonstrate superior performance over ten baseline methods, with significant improvements in both forecasting accuracy and dispatch efficiency.
\end{itemize}

%% file: Chapters/Relatedwork.tex
\vspace{-5pt}
\section{Related Work}
\textbf{Small-scale spatio-temporal prediction models.}
Small-scale spatio-temporal prediction networks are typically composed of temporal and spatial modeling components. Early works, such as STGCN~\cite{yu2018spatio} and ASTGCN~\cite{guo2019attention}, adopt a "sandwich" architecture consisting of a TCN–GCN–TCN structure to capture spatio-temporal dependencies. Recent studies focus on modeling dynamic node correlations. For instance, SHARE~\cite{zhang2020semi} utilizes contextual graph convolution and soft clustering graph convolution, while DyHSL\cite{zhao2023dynamic} models nonpairwise dependencies using hyperedges. For temporal modeling, PDG2seq~\cite{fan2025pdg2seq} decomposes time into daily and weekly features, employing a GRU structure to capture temporal dependencies at different scales. Meanwhile, STGODE~\cite{fang2021spatial} utilizes dilated temporal convolutions to expand its receptive field, thus enhancing its ability to model long-range temporal patterns. These models are often designed for specific scenarios and heavily rely on labeled data, which limits their adaptability and scalability.

\noindent\textbf{LLM and LLM-based models for traffic tasks.}
Large language models have made remarkable progress in recent years. Foundational models such as GPT-4~\cite{achiam2023gpt} and Gemini~\cite{team2023gemini} have laid a solid foundation for the development of intelligent agents based on LLM. Although LLMs were originally developed for text understanding and generation, their adaptability has been widely explored across specialized domains. Researchers have enhanced their ability to model domain-specific knowledge in areas such as code generation~\cite{roziere2023code}, robotic control~\cite{kim2024openvla}, and biomedical information extraction~\cite{luo2022biogpt}. In the traffic domain, UrbanGPT~\cite{li2024urbangpt} improves LLMs’ capacity for temporal modeling by incorporating a time-series encoder. LLMLight~\cite{lai2023llmlight} leverages imitation learning to guide large language models in traffic signal selection. However, these methods are mainly limited to simple regression or classification tasks. They fail to capture complex spatial dependencies and  cannot generalize to more sophisticated planning and decision-making scenarios.

%% file: Chapters/Preliminaries.tex
\vspace{-5pt}

\section{Preliminaries}

We consider two types of urban mobility tasks: (1) spatio-temporal forecasting, which includes both traffic flow and charging demand prediction, and (2) taxi dispatch optimization. For each task type, we describe the construction of the spatial graph, the representation of spatio-temporal data, and the task formulation.

\vspace{-0.5em}
\subsection{Spatio-Temporal Forecasting}

\noindent\textbf{Spatial Structure.}  
The spatial domain is represented as a graph $\mathcal{G} = (\mathcal{V}, \mathcal{E}, \mathcal{A})$, where each node $v_i \in \mathcal{V}$ corresponds to an urban monitoring site, such as a traffic sensor or an urban grid cell. For traffic flow forecasting, edges represent physical road connectivity between sensors, while for charging demand prediction, edges are constructed based on geographical proximity between grid cells. The resulting adjacency matrix $\mathcal{A} \in \mathbb{R}^{N \times N}$ captures spatial relationships among nodes.

\noindent\textbf{Spatio-Temporal Data.}  
Historical observations are encoded as a tensor $\mathbf{X} \in \mathbb{R}^{K \times N \times F}$, where $K$ is the number of past time steps, $N$ is the number of nodes, and $F$ denotes the feature dimension, such as traffic volume, charging demand, or other auxiliary signals.

\noindent\textbf{Task Definition.}  
Given past observations and spatial structure, the goal is to forecast future values over the next $T$ steps for all nodes. Formally:
{\setlength{\belowdisplayskip}{4pt} 
 \setlength{\abovedisplayskip}{4pt} 
\begin{equation}
[\hat{\mathbf{X}}^{t+1}, \ldots, \hat{\mathbf{X}}^{t+T}] = \mathcal{F} \left( 
\mathbf{X}^{[t-K+1:t]}, \mathcal{A} 
\right),
\end{equation}
}
where $\mathcal{F}$ denotes the spatio-temporal forecasting model.

\vspace{-0.5em}
\subsection{Taxi Dispatch Optimization}

\noindent\textbf{Spatial Structure.}  
The urban area is partitioned into a set of grids, each of size 3 km × 3 km, forming the nodes of a graph $\mathcal{G} = (\mathcal{V}, \mathcal{E}, \mathcal{A})$, where each node $v_i \in \mathcal{V}$ represents a region. Edges $\mathcal{E}$ connect spatially adjacent grids, and the adjacency matrix $\mathcal{A} \in \mathbb{R}^{N \times N}$ captures this neighborhood relationship.

\noindent\textbf{Spatio-Temporal Data.}  
At each decision step $t$, we model three key variables for regions $v_c$ together with its eight spatial neighbors, forming a 3×3 neighborhood denoted as $\mathcal{N}_9(c)$.:  
(1) the number of vacant taxis \( \mathbf{X}_{v_c,v}^{t} \in \mathbb{R}^{1 \times 9} \), representing the number of dispatchable vehicles at time \( t \) in the central region \( v_c \) and its eight surrounding neighbors;
(2) the predicted passenger demand \( \mathbf{X}_{v_c,d}^{[t:t+1]} \in \mathbb{R}^{1 \times 9} \), indicating the expected number of ride requests during the interval \( [t, t+1) \);
(3) the predicted number of competing taxis \( \mathbf{X}_{v_c,c}^{[t:t+1]} \in \mathbb{R}^{1 \times 9} \), denoting the number of other available vehicles potentially serving requests.
Notably, we utilize $K$-step historical sequences $\mathbf{X}_{v_c,d}^{[t-K:t-1]}$ and $\mathbf{X}_{v_c,c}^{[t-K:t-1]}$ as inputs to a spatio-temporal prediction encoder, which produces the corresponding future predictions ${X}_{v_c,d}^{[t:t+1]}$ and $\mathbf{X}_{v_c,c}^{[t:t+1]}$.

\begin{figure*}[t]
  \centering
\includegraphics[width=\linewidth]{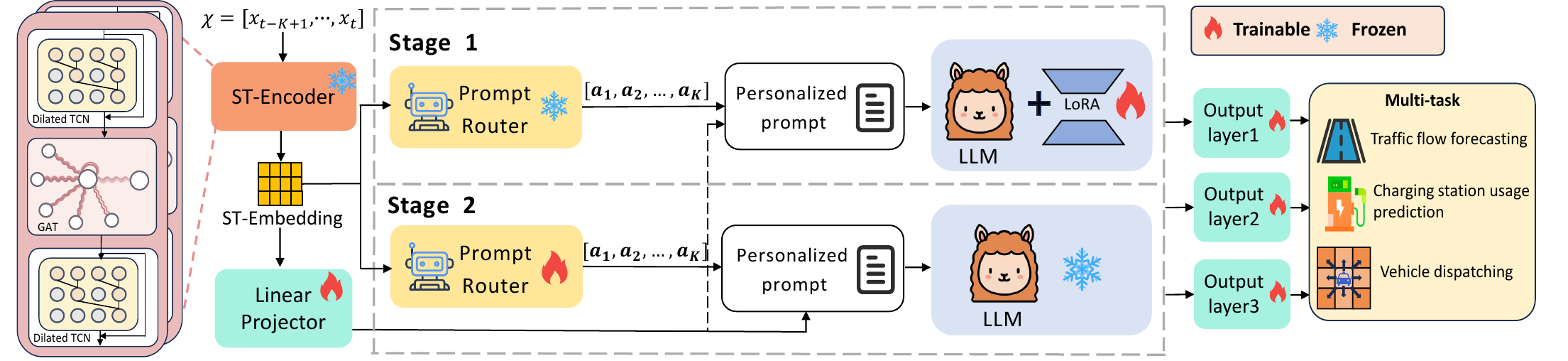}
  \caption{\fontsize{9pt}{\baselineskip}\selectfont The overall architecture of TransLLM.}
  \vspace{-10pt}
    \label{fig1}
\end{figure*}

\noindent\textbf{Task Definition.}  
We formulate the vehicle dispatching task as a localized resource optimization problem, where each decision focuses on reassigning vacant taxis from a central region $v_c$ to its $3 \times 3$ neighborhood $\mathcal{N}_9(c)$. Based on this context, the dispatching decision for region $v_c$ is defined as:
\vspace{-0.3em}
\begin{equation}
\mathbf{D}_c = \mathcal{F} \left( 
\mathbf{X}_{v}^{t_0},\ 
\mathbf{X}_{d}^{[t_0-K:t_0-1]},\ 
\mathbf{X}_{c}^{[t_0-K:t_0-1]};\ 
\mathcal{A}
\right),
\end{equation}
\vspace{-1em}

\noindent where $\mathbf{D}_c \in \mathbb{R}^{1 \times 9}$ denotes the dispatching proportions from $v_c$ each region in $\mathcal{N}_9(c)$.

%% file: Chapters/Method.tex
\vspace{-0.5em}
\section{Method}
In this section, we detail the components of TransLLM, including the architecture of the spatiotemporal encoder (ST-Encoder), the reinforcement learning(RL)-based prompt personalization process, the multi-task output layers, and the overall training mechanism. The overall architecture of TransLLM is illustrated in Figure~\ref{fig1}.

\vspace{-0.5em}
\subsection{Spatio-Temporal Dependency Encoder}


To capture complex spatio-temporal dependencies, we design a spatial-temporal encoder, consisting of multiple spatial-temporal blocks (ST-Blocks). Each ST-Block follows a “sandwich” structure, defined as:
{\setlength{\belowdisplayskip}{4pt} 
 \setlength{\abovedisplayskip}{4pt} 
\begin{equation}
\mathbf{H} = \text{TCN}\left( \text{GAT}\left( \text{TCN}(\mathbf{X}) + \phi(\mathbf{V}), \mathcal{A} \right) \right),
\end{equation}
}
where TCN denotes a dilated Temporal Convolutional Network~\cite{bai2018empirical}, GAT refers to a Graph Attention Network~\cite{velivckovic2017graph} based on the adjacency matrix $\mathcal{A}$, and $\phi(\cdot)$ is an embedding network that encodes node-level meta attributes $\mathbf{V}$. The ST-Block jointly models spatial and temporal dependencies, where dilated TCNs focus on local temporal patterns and GAT learns spatial relationships by attending to neighboring nodes. Besides, we include time-of-day and day-of-week encodings in the input to help the ST-Block capture periodic temporal patterns.

The ST-Encoder employs two distinct adjacency matrices: a spatial adjacency matrix $\mathcal{A}_{sp}$ based on physical road connectivity or geographic proximity, and a semantic adjacency matrix $\mathcal{A}_{se}$ constructed using Dynamic Time Warping~\cite{berndt1994dtw} to capture functional similarity. By incorporating both types of adjacency, the encoder allows each node to aggregate information from not only spatially adjacent regions but also distant regions exhibiting similar temporal behavior. To process these two types of spatial relations, the encoder applies two separate sets of ST-Blocks, producing two spatio-temporal representations: $\mathbf{H}_{sp}$ and $\mathbf{H}_{se} \in \mathbb{R}^{T \times N \times D}$. These are then concatenated to generate the final representation $\mathbf{H_f}$ for downstream predictions.

\vspace{-0.5em}
\subsection{RL-based Prompt Personalization}

\begin{figure}[t]
  \centering
  \includegraphics[width=\linewidth]{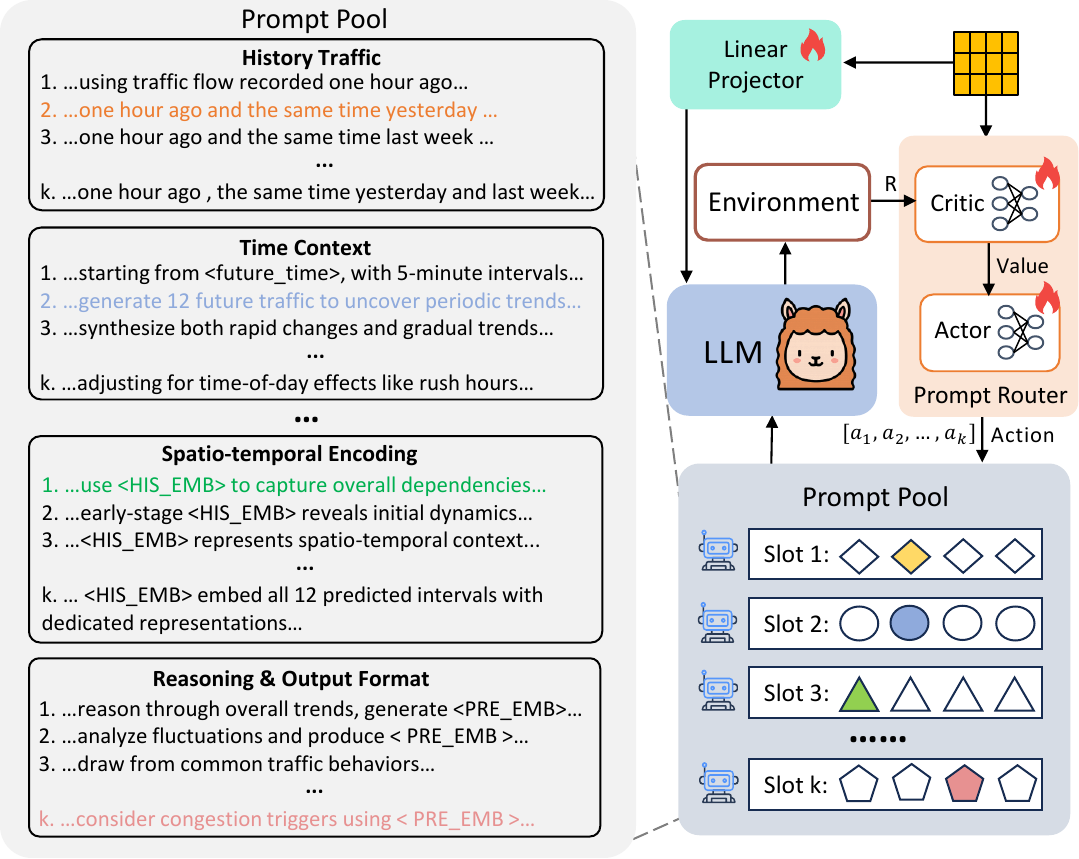}
  \caption{\fontsize{9pt}{\baselineskip}\selectfont Instance-wise prompt generation.}
  \label{fig2}
  \vspace{-10pt}
\end{figure}

To dynamically select the most appropriate prompt for each instance, we introduce a prompt routing mechanism based on an Actor–Critic reinforcement learning framework~\cite{fan2019hybrid}. In this section, we describe the instance-wise prompt generation process, as illustrated in Figure~\ref{fig2}, along with the learning mechanism for updating the router.

\subsubsection{Prompt Routing Mechanism}
We formulate the construction of a personalized prompt as a multi-step decision process. Specifically, a base prompt template is divided into $K$ functional slots, and for each slot, the most suitable component is selected through an independent Actor–Critic network.

To select appropriate sentences for each slot, we use the spatio-temporal representation $\mathbf{H_f}$ generated by the ST-Encoder as input. The actor and critic networks are implemented as two separate MLPs. For each slot $k$, the actor network maps the spatio-temporal representation $\mathbf{H_f}$ to a probability distribution $\pi_k$ over the candidate sentence options, while the critic network computes the expected reward $v_k$ to guide the actor’s training.
The actor and critic functions for slot $k$ are defined as:
\begin{equation}
\begin{aligned}
    \pi_k = \text{Softmax}(\text{MLP}(\mathbf{H_f})), \quad
    V_k = \text{MLP}(\mathbf{H_f}).
\end{aligned}
\end{equation}
An action $a_k$ is sampled from the distribution $\pi_k$, and actions from all $K$ slots are concatenated into a composite vector $\mathbf{a}_t = [a_1, a_2, \ldots, a_K]$, which determines how the final prompt is constructed.

\subsubsection{Diverse Prompt Pool Construction}

The effectiveness of the Prompt Router relies on a well-designed and diverse prompt candidate pool, which constitutes the action space for the RL agent. We construct this pool by partitioning the prompt into $K$ distinct functional slots, each containing multiple candidate sentences that express a similar intent with varied phrasing and focus.
For illustration, in the context of a spatio-temporal forecasting task, we define the following functional slots:

\noindent\textbf{Historical spatio-temporal information:}
To account for the varying periodic influences across different instances, this slot provides historical traffic data at multiple temporal granularities. Examples include "using traffic observations from one hour ago" to capture short-term patterns, or "combining data from the same time on the previous day and week" to incorporate daily and weekly periodic trends. This flexibility enables the model to better adapt to both immediate fluctuations and long-term temporal regularities.

\noindent\textbf{Time Context:} Given the strong periodic patterns in traffic flow, such as morning and evening peaks or holiday effects. This slot specifies the temporal context of the prediction, including the time of day, day of the week, prediction horizon, and temporal resolution. To enhance adaptability across instances, we provide candidate sentences that emphasize different temporal characteristics, such as "capturing short-term fluctuations," or "highlighting daily and weekly periodic trends," encouraging each instance to focus on the most relevant temporal signals for each scenario.

\noindent\textbf{Spatio-temporal Encoding:} To help the language model interpret structured spatiotemporal features, this slot introduces placeholder tokens like $\langle \text{HIS\_EMB} \rangle$, which are replaced by embeddings from a spatio-temporal encoder. The candidate sentences vary in focus—some describe full 12-step prediction dependencies, others emphasize early-stage forecasting or the interpretability of step-wise representations. These variations allow the model to flexibly adjust its reasoning strategy based on the temporal characteristics of each instance.

\noindent\textbf{Reasoning \& Output Format:} This slot defines how the language model should generate its prediction and present the output. While all candidate sentences guide the model to conclude with a standardized 12-step forecast using the $\langle \text{PRE\_EMB} \rangle$ token, they differ in the reasoning process they encourage. Some variants explicitly instruct the model to reason step by step through temporal patterns and spatial influences, aiming to enhance interpretability and reduce hallucination. Others emphasize domain knowledge—such as rush hour behavior or anomalies in recent flow trends—to guide the forecast more heuristically. This diversity allows the model to flexibly adapt its reasoning strategy to suit different traffic scenarios.

\subsubsection{Textual-Spatiotemporal Alignment.} 

To enable effective alignment between structured spatiotemporal signals and natural language input, we embed the encoded representations directly into the LLM's input space via a token-based interface. Specifically, we use a placeholder token sequence $\langle \text{st\_start} \rangle\, \langle \text{st\_patch} \rangle^{\times T}\, \langle \text{st\_end} \rangle$ to replace $\langle \text{HIS\_EMB} \rangle$ and $\langle \text{PRE\_EMB} \rangle$, where $T$ denotes the number of prediction steps, and each $\langle \text{st\_patch} \rangle$ corresponds to one future time step.
Subsequently, the spatio-temporal embeddings $\mathbf{E} \in \mathbb{R}^{T \times 2D}$ produced by the ST-Encoder are projected to match the LLM’s hidden dimension $d_{\text{L}}$ via a linear layer, 
yielding $\mathbf{E}' \in \mathbb{R}^{T \times d_{\text{L}}}$. These aligned embeddings are then inserted into the prompt by sequentially replacing 
the $\langle \text{st\_patch} \rangle$ tokens.

\subsubsection{Reinforcement-Guided Routing Update}
To support diverse downstream tasks, the reward signal $\hat{R}$ for training the prompt router is defined as either the negative prediction loss or task-specific metrics such as taxi dispatch rewards, directly linking prompt quality to task performance. For each router, we apply a shared reward signal to evaluate all slots, enabling each slot to independently learn its optimal prompt routing policy through dedicated actor-critic networks. The optimization objectives for the actor and critic networks of slot $k$ are defined as follows:
\begin{equation}
\begin{aligned}
\mathcal{L}_a^{(k)} = -\log \pi_k\cdot \left( \hat{R}_{t} - V_k \right),\quad
\mathcal{L}_c^{(k)} =  \left( \hat{R}_{t} - V_k \right)^2.
\end{aligned}
\end{equation}

\subsection{Multi-task Output Layers}

In TransLLM, we avoid using the LLM to directly generate outputs. Instead, we employ multiple task-specific output layers. This design enables more accurate handling of continuous values, avoiding the precision loss caused by token-level discretization. Moreover, customized output layers are better aligned with the objective formats and evaluation metrics of different downstream tasks, helping to mitigate interference caused by mismatched output spaces across tasks. The hidden representations from the LLM are extracted and passed through a multi-task output layer tailored to produce task-specific predictions.

\noindent\textbf{Spatio-Temporal Forecasting:} The prediction-related hidden state $\mathbf{H_f}'$ at the $\langle \text{st\_start} \rangle$ token and the encoder output $\mathbf{H_f}$ are individually processed by linear output layers to generate the final prediction $\hat{\mathbf{y}}$:
\vspace{-2pt}
\begin{equation}
\small
\hat{\mathbf{y}} = \mathbf{W}_3 \left[ \text{Concat} \left( \text{ReLU}(\mathbf{W}_1 \mathbf{H_f} + \mathbf{b}_1), \; \text{ReLU}(\mathbf{W}_2 \mathbf{H_f}' + \mathbf{b}_2) \right) \right] + \mathbf{b}_3,
\end{equation}
\vspace{-12pt}

\noindent where $\mathbf{W}_1$, $\mathbf{W}_2$, $\mathbf{W}_3$, $\mathbf{b}_1$, $\mathbf{b}_2$, and $\mathbf{b}_3$ are learnable weight matrices.

\noindent\textbf{Taxi Dispatch Optimization:}  
For the dispatching task, the goal is to generate a probability distribution over the nine candidate regions in the 3×3 neighborhood surrounding the current location. To this end, the hidden state $\mathbf{H}_{\text{f}}'$ from the $\langle \text{st\_start} \rangle$ token is first projected through a linear layer and then normalized via a softmax function to produce the dispatching probability distribution:
\vspace{-2pt}
\begin{equation}
\mathbf{D}_c = \text{Softmax}(\mathbf{W}_d \mathbf{H}_{\text{f}}' + \mathbf{b}_d),
\end{equation}
\vspace{-12pt}

\noindent where $\mathbf{W}_d$ and $\mathbf{b}_d$ are learnable parameters. The output $\mathbf{D}_c \in \mathbb{R}^{1 \times 9}$ indicates the proportion of vacant vehicles that should be reallocated from grid $v_c$ to its surrounding regions.


\vspace{-0.5em}
\subsection{Model Training}
\subsubsection{Two-stage Alternating Training }

 Considering the high computational cost of full-parameter fine-tuning for LLMs, we adopt Low-Rank Adaptation (LoRA)\cite{hu2022lora} to improve training efficiency. To improve stability and efficiency, we adopt a two-stage training strategy. In the first stage, the LLM is fine-tuned, while the Prompt Router remains frozen. In the second stage, the LLM is frozen and the Prompt Router is trained. In practice, we find that a single alternation between the two stages is sufficient to reach optimal performance on traffic-related tasks.

\subsubsection{Loss Function}
\label{subsec:Model Optimization}

To enable joint optimization across multiple tasks, we design a composite loss function formulated as:
\vspace{-0.5em}
\begin{equation}
\mathcal{L} = \mathcal{L}_{LLM} + \lambda_{t} \cdot \mathcal{L}_{t},
\end{equation}
\vspace{-1.5em}

\noindent where $\mathcal{L}_{LLM}$ denotes the standard cross-entropy loss for LLM, $\mathcal{L}_{t}$ represents the specific task loss, and $\lambda_{t}$ is a weighting coefficient.

\noindent\textbf{Spatio-Temporal Forecasting Loss.}
We employ the Mean Absolute Error (MAE) loss as the regression loss:
\vspace{-0.8em}
\begin{equation}
\mathcal{L}_{t} = \frac{1}{N}\sum_{i=1}^{N} | \hat{y}_{i} - y_{i} |,
\end{equation}
\vspace{-1.5em}

\noindent where $\hat{y}_{i}$ and $y_{i}$ denote the predicted and ground truth values for the $i$-th sample.

\noindent\textbf{Reinforcement Learning Loss for Vehicle Dispatching.}
We propose a multi-objective reinforcement learning loss function:
\vspace{-0.5em}
\begin{equation}
\begin{aligned}
\mathcal{L}_{d} &= \mathcal{L}_{rf}
            + \lambda_{w}\,\mathcal{L}_{w} + \lambda_{e}\,\mathcal{H}(\pi),\quad
\mathcal{H}(\pi) = -\sum_{g=1}^{9} \pi_{g} \log \pi_{g},\\
\mathcal{L}_{w} &=\sum_{g=1}^{9} \left| \text{CDF}(\pi_b)_g - \text{CDF}(\mathbf{p}_{\text{real}}^{b})_g \right|,\quad 
R_g = \beta\,
      \frac{M_g}{V_g}
      - \gamma D_g.
\end{aligned}
\end{equation}
\vspace{-1.5em}

\noindent where $\mathcal{L}_{rf}$ is the reinforcement loss, $\mathcal{L}_{w}$ is the Wasserstein distance between predicted and ground-truth distributions computed via cumulative distribution functions (CDFs), $\mathcal{H}(\pi)$ is the entropy of the predicted distribution, and $R_g$ is the reward function for reinforcement learning that balances service efficiency and dispatch cost. Specifically, $M_g$ is the matching rate in grid $g$, $V_g$ is the number of vacant taxis, and $D_g$ represents a predefined distance penalty. The coefficients $\lambda_w$, $\lambda_e$, $\beta$, and $\gamma$ are the weighting factors for each component. This composite loss structure enables our model to learn a sophisticated dispatching policy that simultaneously balances service efficiency, dispatching cost, and distributional realism.

%% file: Chapters/Experiment.tex
\vspace{-0.5em}
\section{Experiment}
In this section, we conduct extensive experiments to evaluate the effectiveness of TransLLM across multiple transportation tasks. Our study is guided by the following research questions. Additionally, a case study is provided in the Appendix.
\begin{itemize}[leftmargin=*]
    \item \textbf{RQ1:} How does TransLLM perform compared to GNN-based baselines in traditional supervised scenarios? Can it effectively handle complex real-world taxi dispatching task?

    \item \textbf{RQ2:} How is the generalization ability of TransLLM when facing unseen scenarios and datasets? 

    \item \textbf{RQ3:} What is the contribution of each key module within TransLLM to its overall performance? 

    \item \textbf{RQ4:} How does the performance of TransLLM vary under different hyperparameter configurations settings
\end{itemize}

\vspace{-1em}
\subsection{Dataset}
To comprehensively evaluate the effectiveness of TransLLM across diverse transportation tasks, we conduct experiments on five public datasets covering traffic forecasting, charging demand prediction, and vehicle dispatching. LargeST-SD is a subset of the LargeST dataset~\cite{liu2023largest}, containing traffic flow records from 716 loop detectors on highways in San Diego County, spanning from January 1 to December 31, 2021. Pems08 includes traffic data from 170 sensors in California, collected between July 1 and August 31, 2016. ST-EVCDP~\cite{qu2024physics} is a charging demand dataset collected from 247 areas in Shenzhen between June 19 and July 18, 2022, while UrbanEV~\cite{li2025urbanev} captures large-scale charging behavior across urban regions from September 1, 2022 to February 28, 2023. For the vehicle dispatching task, we use the Taxi-SH dataset, which contains GPS trajectories of taxis in Shanghai from April 13 to April 19, 2015. The city is divided into 3 km × 3 km grids, and we aggregate the number of vacant taxis, passenger demand, and competing vehicles at 5-minute intervals. In addition, we evaluate the models under a zero-shot setting on two previously unseen datasets, PEMS03 and PEMS04, to assess their generalization capabilities. Among these, UrbanEV has a temporal resolution of one hour, while the others have a resolution of five minutes. More details are provided in the Appendix.

\subsection{Hyperparameters Settings}
In our experimental setup, the key hyperparameters are configured as follows. The ST-Encoder is built by stacking 2 ST-Blocks, with a final output dimension $D$ of 64. The Prompt Router is configured with 4 slots and 4 sentences. For prediction tasks, we predict the next 12 steps of data based on the previous 12 steps. Both the history length $K$ and prediction length $T$ are set to 12. For the taxi dispatch task, we predict the dispatch probabilities over nine grids. Both $K$ and $T$ are set to 9. The model is trained with a learning rate of $1 \times 10^{-4}$, and the composite loss function balances objectives with weights $\lambda_t=1.0$, $\lambda_w=0.01$, $\lambda_e=0.008$, and reward coefficients $\beta=2.0$ and $\gamma=0.05$.

\vspace{-0.5em}
\subsection{Evaluation Metrics}

For spatio-temporal forecasting tasks, we employ the Mean Absolute Error (MAE) and the Root Mean Squared Error (RMSE) to quantify prediction accuracy. Both metrics measure the deviation between predicted and true values.
\begin{equation}
\label{eq:mae}
\text{MAE} = \frac{1}{N} \sum_{i=1}^N |y_i - \hat{y}_i|,\quad
\text{RMSE} = \sqrt{\frac{1}{N} \sum_{i=1}^N (y_i - \hat{y}_i)^2}.
\end{equation} 
\vspace{-0.5em}

For the taxi dispatching task, we introduce three evaluation metrics: the Mean Matching Rate (MMR) represents the proportion of matched vehicles; the Mean Driving Distance (MDD) calculates the average deadhead distance to measure dispatching costs; and the Wasserstein Distance (W-Dist) quantifies the discrepancy between the model's predicted dispatch distribution and the ground-truth distribution.The three metrics are formulated as:
\begin{equation}
\begin{aligned}
&\text{MMR} = \frac{1}{N} \sum_{i=1}^{N} \left( \frac{\sum_{g=1}^9 \hat{\text{M}}_{i,g}}{\text{V}_{0,i}} \right),
\text{MDD} = \frac{1}{N} \sum_{i=1}^{N} \left( \sum_{g=1}^9 \hat{p}_{i,g} \cdot C_k \right),\\
&\text{W-Dist} = \frac{1}{N} \sum_{i=1}^{N} \left( \sum_{g=1}^9 |CDF(\hat{P}_{i})_g - CDF(P_{i})_g| \right).
\end{aligned}
\end{equation}
\vspace{-1em}

\noindent Here, $\hat{\text{M}}_{i,g}$ is the predicted matched vehicles in region $g$ for instance $i$; $\text{V}_{0,i}$ represents the empty vehicles in the central region; $\hat{p}_{i,g}$ is the predicted dispatch probability; $C_g$ denotes the deadhead distance cost; and $\hat{P}_{i}$ and $P_{i}$ are the predicted and true dispatch probability distributions.
\vspace{-0.5em}
\subsection{Baselines}
We conducted a thorough comparison with ten baseline models to ensure a comprehensive evaluation of our proposed approach. These baselines can be broadly categorized into three groups:
\begin{itemize}[leftmargin=*]
\item \textbf{Small-scale GNN-based Deep Learning Models.}
These models are developed specifically for spatiotemporal forecasting and are trained in an end-to-end manner on individual datasets. They typically rely on carefully designed architectures that integrate graph-based spatial encoders with temporal modeling components. Representative examples include STGCN~\cite{yu2018spatio}, DGCRN~\cite{li2023dynamic}, ASTGCN~\cite{guo2019attention}, D$^{2}$STGNN~\cite{shao2022decoupled}, GWNET~\cite{wu2019graph}, STGODE~\cite{fang2021spatial}, and PDG2Seq~\cite{fan2025pdg2seq}.
\item \textbf{Generalist Large Language Models (LLMs).}
These models, including Deepseek-v3~\cite{liu2024deepseek} and GPT-4o~\cite{hurst2024gpt}, are not specifically trained for spatiotemporal or transportation tasks. We evaluate them in a zero-shot setting without any task-specific tuning:
\item \textbf{Spatio-temporal Enhanced LLMs.}
This category includes LLM-based architectures that leverage structured spatiotemporal priors to improve the model’s ability to reason over urban dynamics, with UrbanGPT~\cite{li2024urbangpt} as a representative example.
\end{itemize}
\vspace{-1em}
\subsection{Overall Performance (RQ1)}
\begin{table*}[t]
    \centering
    \caption{\fontsize{9pt}{\baselineskip}\selectfont Performance comparison across model types. \textbf{Bold}: Best, \underline{underline}: Second best, \doubleunderline{Double underline}: Third best.}
    \small
    \setlength{\tabcolsep}{7pt}
    \begin{tabular}{@{}c|cc|cc|cc|cc|ccc@{}}
        \toprule
          & \multicolumn{4}{c}{Traffic Forecasting} & \multicolumn{4}{c}{Charging Demand Prediction} & \multicolumn{3}{c}{Vehicle Dispatching} \\
          \cmidrule(lr){2-5}\cmidrule(lr){6-9}\cmidrule(lr){10-12}
         Models & \multicolumn{2}{c}{LargeST-SD } & \multicolumn{2}{c}{PEMS08} &\multicolumn{2}{c}{ST-EVCDP } & \multicolumn{2}{c}{UrbanEV} &\multicolumn{3}{c}{Taxi-SH} \\
         \cmidrule(lr){2-3}\cmidrule(lr){4-5}\cmidrule(lr){6-7}\cmidrule(lr){8-9}\cmidrule(lr){10-12}
        & MAE~$\downarrow$ & RMSE~$\downarrow$ & MAE~$\downarrow$ & RMSE~$\downarrow$ 
        & MAE~$\downarrow$ & RMSE~$\downarrow$ & MAE~$\downarrow$ & RMSE~$\downarrow$ 
        & MMR(\%)~$\uparrow$ & MDD~$\downarrow$ & W-Dist~$\downarrow$ \\

        \midrule
        \multicolumn{12}{c}{\textbf{Small-scale GNN-based Deep Learning Models}} \\
        \midrule
        STGCN &13.93 &26.10  &10.30&14.77& 2.11   &3.64  &3.02&5.30   & 19.10    & 5.80   &1.26      \\
         DGCRN    & 11.46 & 24.63 &10.45&14.47& 1.71    & 4.46 &3.04&5.47   & 18.89    & 3.60    & 1.46     \\
         ASTGCN    & 12.34 & 25.09 &10.17&14.48& 1.85    & 3.45   &3.32&5.81& \doubleunderline{19.91}    & 3.96    & 1.32     \\
         D$^{2}$STGNN   & 11.72 & 25.18 &\doubleunderline{8.89}&\doubleunderline{12.72}& \underline{1.39}    & \underline{2.56}    &2.65 &4.59& 18.66  & 3.50    & 1.25     \\
         GWNET     & 13.43 & 26.42 &9.26&12.90& 1.45    & \doubleunderline{2.65}   &2.57&\doubleunderline{4.27} & 19.29    & 3.49    & 1.34     \\
         STGODE    & 11.83 & 24.63 &9.18&13.16& 1.50    & 2.73   &2.66&4.53 & 19.26    & 3.47    & 1.28     \\
         PDG2seq   & 12.16    & 25.02   &9.54&13.53 & 1.46    & 2.93   &3.33&7.38 & 18.92    & 3.49    & \doubleunderline{1.21}     \\
        \midrule
        \multicolumn{12}{c}{\textbf{Generalist LLMs}} \\
        \midrule
         Deepseek-v3 & 39.25    & 52.28   &23.41& 30.26& 2.13    & 4.73    &4.60&9.02& 19.07    & \underline{2.57}    & \textbf{0.75}    \\
         GPT-4o    & 40.54    & 53.52   &23.89& 30.89& 1.66 &3.55   &3.53&7.85 &18.00    & \textbf{2.40}    & \underline{0.86}     \\
        \midrule
        \multicolumn{12}{c}{\textbf{Spatio-temporal Enhanced LLMs}} \\
        \midrule
         Urbangpt & \doubleunderline{11.28} & \doubleunderline{23.17} &10.23&13.53& 2.09 & 5.86 &\doubleunderline{1.12} &4.38& 19.46 & 3.21 & 2.31 \\
         TransLLM(vicuna) & \underline{10.98} & \underline{21.42} &\underline{7.88}&\underline{11.33}&\underline{1.39} & 2.73 &\underline{0.44}&\underline{3.64} & \underline{24.46} & \doubleunderline{3.15} & 1.96 \\
         TransLLM(llama3) & \textbf{9.41} & \textbf{16.78} & \textbf{7.26}&\textbf{10.68}&\textbf{1.26} & \textbf{2.16} &\textbf{0.42} &\textbf{3.42}&\textbf{24.78} & 3.25 & 2.02 \\
        \bottomrule
    \end{tabular}
    \label{tab:overall_performance}
\end{table*}

\begin{table}[t]
    \centering
    \caption{\fontsize{9pt}{\baselineskip}\selectfont Zero-shot performance on PEMS03 and PEMS04.}
    \small
    \setlength{\tabcolsep}{10pt}
    \begin{tabular}{@{}c|cc|cc@{}}
        \toprule
         & \multicolumn{2}{c}{PEMS03} & \multicolumn{2}{c}{PEMS04} \\
         \cmidrule(lr){2-3}\cmidrule(lr){4-5}
         Models & MAE~$\downarrow$ & RMSE~$\downarrow$ & MAE~$\downarrow$ & RMSE~$\downarrow$ \\
        \midrule
        \multicolumn{5}{c}{\textbf{Small-scale GNN-based Deep Learning Models}} \\
        \midrule
        STGCN & 57.43 & 64.74 & 39.08 & 49.93 \\
        DGCRN & 58.05 & 61.86 & \doubleunderline{29.60} & \doubleunderline{39.30} \\
        ASTGCN & 80.23 & 88.66 & 45.02 & 54.65 \\
        D$^{2}$STGNN & 63.17 & 66.65 & 30.92 & \underline{39.23} \\
        GWNET & 55.51 & 59.21 & 
        \textbf{27.06} & \textbf{33.46} \\
        STGODE & 73.54 & 81.98 & 64.33 & 77.18 \\
        PDG2seq & 88.07 & 101.10 & 55.88 & 69.15 \\
        \midrule
        \multicolumn{5}{c}{\textbf{Generalist LLMs}} \\
        \midrule
        Deepseek-v3 & \doubleunderline{25.15} & \doubleunderline{38.28} & 37.79 & 55.83 \\
         GPT-4o & 27.06 & 41.31 & 40.42 & 59.47 \\
        \midrule
        \multicolumn{5}{c}{\textbf{Spatio-temporal Enhanced LLMs}} \\
        \midrule
        Urbangpt & \underline{23.27} &  \underline{37.58} & 41.52 & 69.76 \\
        TransLLM & \textbf{18.92} & \textbf{30.27} & \underline{28.69} & 47.27 \\
        \bottomrule
    \end{tabular}
    \label{tab:zeroshot_performance}
    \vspace{-2em}
\end{table}

We evaluate the performance of TransLLM across three tasks and five datasets, including traffic flow forecasting, charging demand prediction, and taxi dispatch optimization. Table~\ref{tab:overall_performance} summarizes the overall results. TransLLM (LLaMA3) and TransLLM (Vicuna) refer to model variants built on different base LLMs. Overall, even in scenarios where small-scale models typically excel, TransLLM consistently outperforms all baselines across the five datasets. Moreover, employing more advanced foundation language models can lead to further improvements in the performance of TransLLM.

We observe that the performance of the three model categories varies across different datasets. On the UrbanEV dataset, spatio-temporal enhanced LLMs clearly outperform the other two categories. In particular, TransLLM achieves a MAE of 0.42, representing an 83.7\% improvement over the best-performing small-scale baseline. This may be attributed to the dataset's 1-hour temporal resolution, which contrasts with the 5-minute intervals used in other datasets. The smoother data distribution weakens the advantage of small models in capturing fine-grained short-term dependencies. In contrast, LLM-based models excel by leveraging contextual reasoning and learned priors, which enhance their robustness to sparse temporal signals. Meanwhile, the performance gain of TransLLM on the ST-EVCDP dataset is notably smaller than that on traffic flow prediction datasets such as PEMS08. To investigate this discrepancy, we found that ST-EVCDP exhibits weaker periodic patterns and higher volatility compared to PEMS08, as detailed in the Appendix, making it inherently more difficult to predict. Moreover, TransLLM is trained with significantly less data than small-scale models, which makes achieving high accuracy more challenging. In contrast, powerful foundation models such as GPT-4o demonstrate strong performance on ST-EVCDP, likely benefiting from their rich and diverse pretraining data.

The effectiveness of TransLLM is further validated on the taxi dispatching task. The quality of dispatch is evaluated using three metrics: MMR, MDD, and W-Dist to the real distribution. Among these, MMR is the most critical metric. Compared to the second-best model, TransLLM improves MMR by 5\% and reduces the average travel distance by 0.71 km. TransLLM underperforms GPT-4o and DeepSeek-v3 in terms of MMD and W-Dist. This is because generalist LLMs tend to adopt a conservative dispatching strategy, keeping vehicles in their original grids, which leads to the lowest average travel distance and the closest match to the real-world distribution.

In summary, despite limited training data, TransLLM consistently outperforms ten baseline models in spatiotemporal forecasting tasks. For the taxi dispatching task, TransLLM adopts a more proactive dispatching strategy that achieves higher order-taking rates with only a slight increase in travel distance, despite deviating from real trajectories.

\vspace{-0.5em}
\subsection{Zero-shot Scenarios Performance(RQ2)}



Zero-shot generalization is critical for foundation models in the transportation domain. It reflects the model’s ability to capture universal spatiotemporal patterns and determines its transferability and practical value across regions and tasks. A model with strong generalization capabilities can adapt to prediction tasks in new cities and unseen scenarios without the need for additional annotations or fine-tuning, thereby significantly reducing deployment costs and enhancing the feasibility of large-scale applications. Specifically, we utilize datasets from different districts within the Los Angeles area: PEMS08 is used for training, while zero-shot evaluation is conducted on PEMS03 and PEMS04, both of which remain unseen during the training phase.

We observe that LLM-based methods achieve significantly better improvements over small-scale GNN-based deep learning models on the PEMS03 dataset. In contrast, on the PEMS04 dataset, both categories exhibit similar performance. This discrepancy may be due to PEMS04 having more similar spatiotemporal patterns to the training dataset, PEMS08, than PEMS03. Specifically, the PEMS08 dataset has an average flow of 230.68 and a maximum of 1147.0, while PEMS04 has an average flow of 226.13 and a maximum of 896.0. In comparison, PEMS03 features a lower average flow of 149.52 but a much higher peak of 1852.0. These more similar traffic patterns and the limited training data reduce the relative advantages of LLM-based methods.
Nevertheless, TransLLM still achieves the best predictive performance among all LLM-based approaches. On the more divergent PEMS03 dataset, it outperforms UrbanGPT by 18.7\% in MAE and 19.5\% in RMSE.

\subsection{Ablation Study (RQ3)}

\begin{figure}[t]
  \centering
\includegraphics[width=\linewidth]{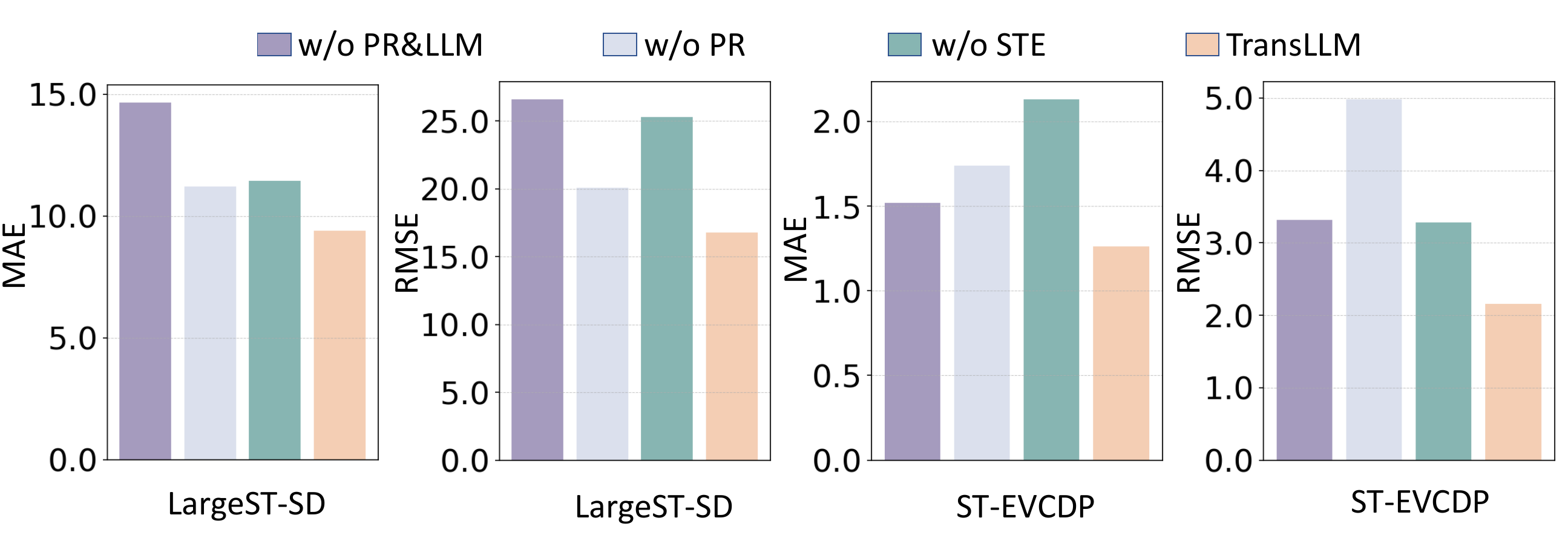}
  \caption{\fontsize{9pt}{\baselineskip}\selectfont Ablation study results on LargeST-SD and ST-EVCDP datasets.}
  \vspace{-5pt}
    \label{fig3}
\end{figure}

To investigate the impact of individual modules on the overall performance, we evaluated TransLLM variants by removing key components: the ST-Encoder, Prompt Router, and the LLM module. Figure~\ref{fig3} presents the ablation results on the LargeST-SD and ST-EVCDP datasets.

\noindent\textbf{Effect of ST-encoder.} w/o STE denotes a variant of TransLLM in which the ST-Encoder modules are removed. We observe a noticeable decline in the model’s performance, indicating that the rich spatio-temporal features captured by the ST-Encoder effectively assist the LLM in understanding and processing complex spatio-temporal patterns. The ST-Encoder serves as a crucial component that integrates non-textual spatio-temporal information into the LLM’s input space, and its removal results in a substantial degradation in predictive accuracy.

\begin{figure}[t]
  \centering
\includegraphics[width=\linewidth]{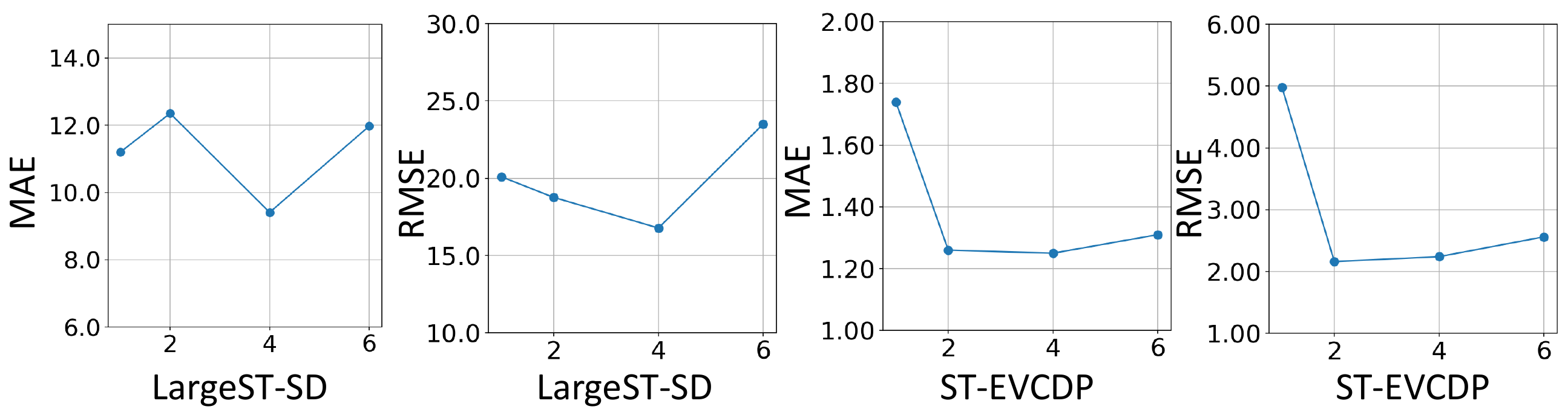}
  \caption{\fontsize{9pt}{\baselineskip}\selectfont Effect of number of candidate sentences per slot $N_c$ on LargeST-SD and ST-EVCDP datasets.}
  \vspace{-15pt}
    \label{fig4}
\end{figure}

\noindent\textbf{Effect of Prompter Router.} w/o PR removes the Prompt Router module and instead uses the same fixed prompt for each instance. This variant exhibits a clear performance drop. Although the LLM can still leverage the spatio-temporal features provided by the ST-Encoder, the absence of dynamically generated, instance-specific prompts limits its full potential. This highlights the critical role that prompt quality plays in LLM performance. However, prompt quality is inherently difficult to evaluate manually. The Prompt Router plays a vital role in supplying the LLM with accurate and instance-specific prompts, thereby enhancing its reasoning and prediction capabilities.


\noindent\textbf{Effect of Large language model.} w/o PR \& LLM removes the Prompt Router and the LLM modules, also resulting in a significant performance decline. This underscores the LLM’s ability to model complex spatio-temporal patterns, support advanced reasoning, and enhance generalization. 
To summary, the results demonstrate that the full model consistently achieves the best performance across all tasks, and the absence of any individual module leads to a degradation in TransLLM’s overall effectiveness.

\vspace{-0.5em}
\subsection{Parameter Sensitivity (RQ4)}


To evaluate the impact of different hyperparameters on TransLLM’s performance, we adjust two key settings: (1) the number of candidate sentences per slot $N_c$, and (2) the number of spatio-temporal feature patch tokens $\langle \text{st\_patch} \rangle$ $N_p$.

\noindent \textbf{Number of candidate sentences per slot $N_c$.} Figure~\ref{fig4} shows that increasing the number of sentences does not necessarily lead to continuous performance improvements. The optimal configuration is achieved with four candidate sentences per slot on LargeST-SD and two on ST-EVCDP. While an insufficient number of sentences may limit the diversity available to the Prompt Router, an excessive number can introduce redundancy and increase the learning burden on the LLM, ultimately degrading performance.

\begin{figure}[t]
  \centering
\includegraphics[width=\linewidth]{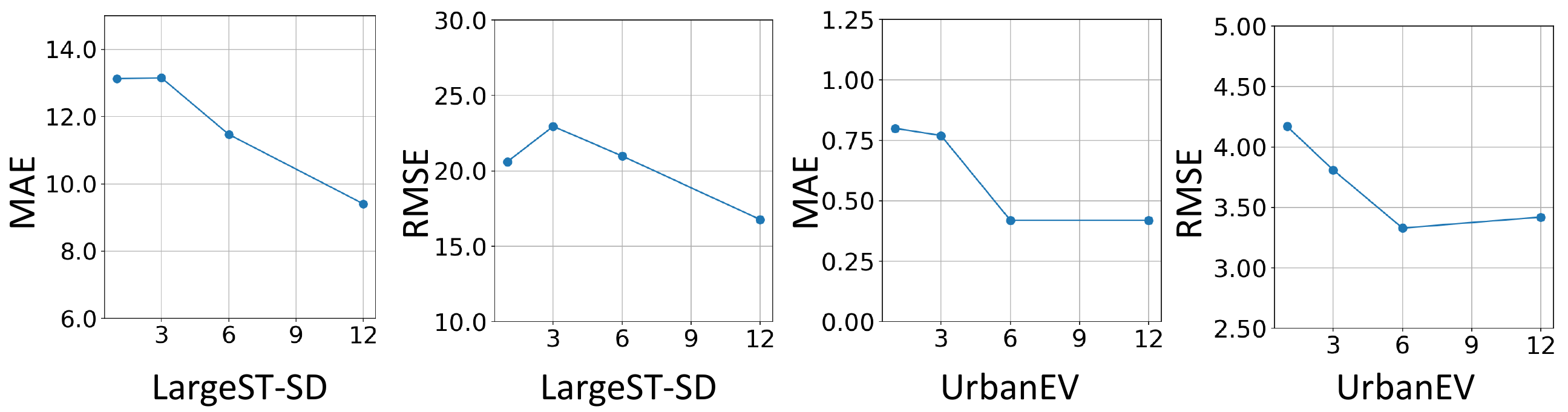}
  \caption{\fontsize{9pt}{\baselineskip}\selectfont Effect of number of $\langle \text{st\_patch} \rangle$ tokens $N_p$ on LargeST-SD and UrbanEV datasets.}
  \vspace{-15pt}
    \label{fig5}
\end{figure}

\noindent \textbf{Number of $\langle \text{st\_patch} \rangle$ tokens $N_p$.}
Figure~\ref{fig5} indicates that model performance generally improves with an increasing number of tokens. 
The best results are typically obtained when the token count matches the prediction horizon. This alignment enables comprehensive encoding of spatio-temporal dependencies, whereas shorter sequences such as 1 or 3 tokens fail to capture sufficient information, leading to degraded performance. Notably, on the UrbanEV dataset, both 6 and 12 tokens yield similarly strong results, likely due to its lower temporal resolution, which reduces sensitivity to token sequence length. In summary, the above parameter sensitivity analysis provides important insights for selecting TransLLM's hyperparameters, ensuring the model's robustness and optimal performance across different tasks and datasets.

\vspace{-0.5em}
\subsection{Prompt Routing Behavior Analysis }
\begin{figure}[t]
  \centering
\includegraphics[width=\linewidth]{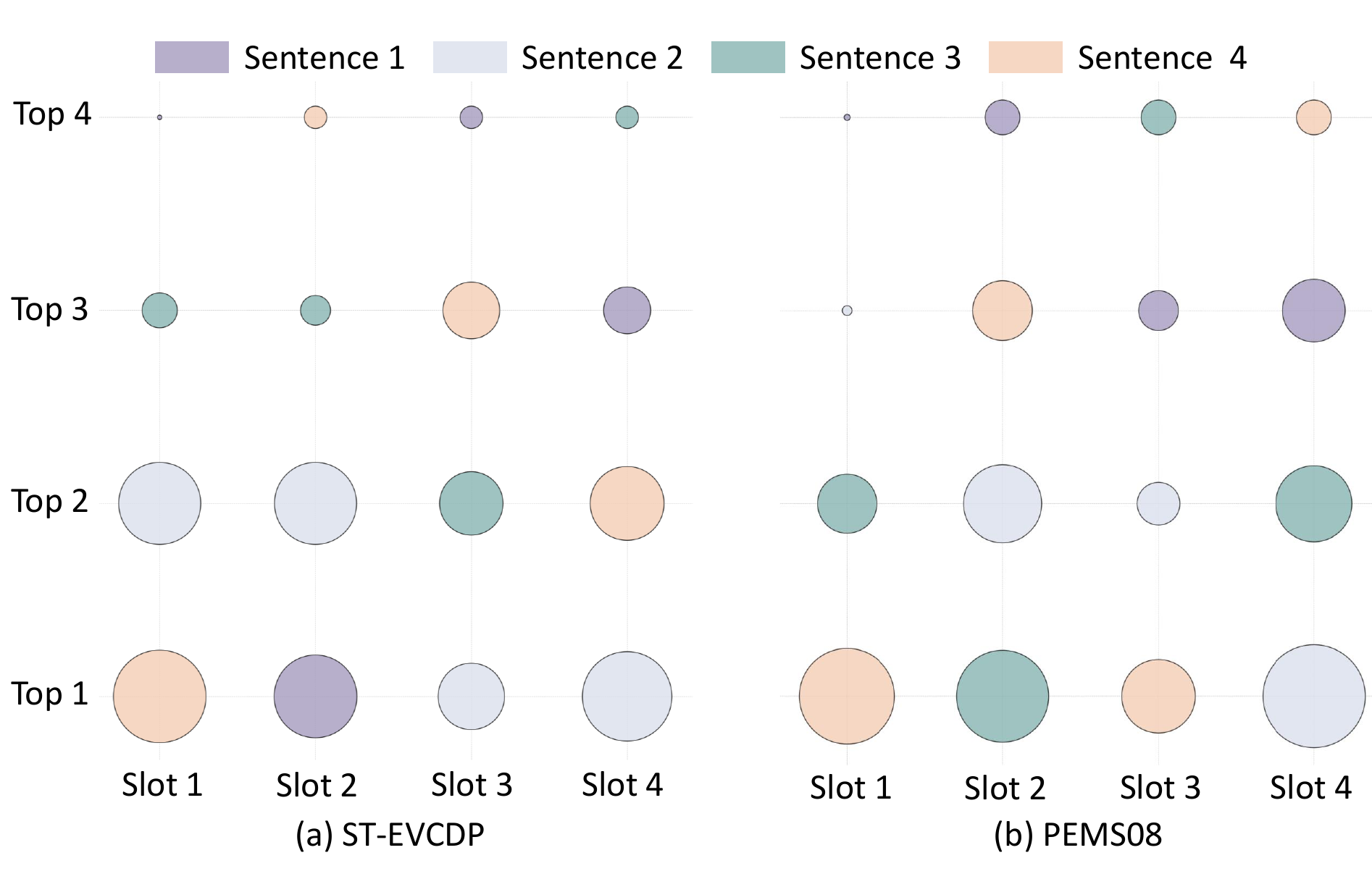}
  \caption{\fontsize{9pt}{\baselineskip}\selectfont Sentence selection frequency per slot. The size of the circles
represents the selection frequency of each sentence.}
  \vspace{-15pt}
    \label{fig6}
\end{figure}

To further evaluate the effectiveness of the Prompt Router, we visualize the selection frequency of each sentence in the prompt pool during testing on the ST-EVCDP and PEMS08 datasets, as shown in Figure~\ref{fig6}. It can be observed that the Prompt Router successfully selects the most appropriate prompt for each instance. Moreover, the selection frequencies vary across sentences within each slot—some sentences are favored by the majority of instances, while others are rarely chosen. This also reveals the diversity of spatiotemporal patterns across instances and the router's ability to adaptively match prompts based on instance-level characteristics.

%% file: Chapters/Conclusion.tex
\vspace{-0.5em}
\section{Conclusion}
In this work, we present TransLLM, a unified foundation framework that seamlessly integrates spatiotemporal modeling with large language models for diverse urban transportation tasks. Our key contributions include the novel framework design that bridges spatiotemporal encoders with LLMs through structured embeddings, and a two-stage learnable prompt composition mechanism with instance-level routing that dynamically personalizes prompts based on input characteristics. Extensive experiments across seven datasets demonstrate strong performance in both supervised and zero-shot settings, showcasing excellent generalization and cross-task adaptability.

%% file: Chapters/Appendix.tex
\appendix

\vspace{-0.5em}
\section{Appendix}

\subsection{Ethical Use of Data, Informed Consent, and Dataset Details} 

This study is based entirely on open-access datasets curated for research applications. It does not involve human subject interaction, nor does it process any personal or sensitive information. All data handling respects the usage terms of the original providers and aligns with the ethical research standards outlined by ACM.
  
We train both TransLLM and the baseline models on five datasets, covering three types of tasks: traffic forecasting, charging demand prediction, and vehicle dispatching. Specifically, Large-SD and PEMS08 are used for traffic forecasting, ST-EVCDP and UrbanEV for charging demand prediction, and Taxi-SH for vehicle dispatching. Among these, UrbanEV has a temporal resolution of one hour, while the others have a resolution of five minutes.

We perform temporal splits for each dataset. For all small-scale baseline models, the first 30\% of each dataset is used for training. Due to the higher computational cost of training TransLLM, we employ a reduced training subset. For instance, only 4 days of data from Large-SD and 24 days from UrbanEV are used to train TransLLM.  
For evaluation, we extract a test set of $N \times 12$ samples from each dataset, where $N$ is the number of nodes, ensuring that all samples are drawn from unseen time periods. To assess generalization under zero-shot settings, we evaluate on two previously unseen datasets, PEMS03 and PEMS04, by extracting 2 hours of data from 170 nodes in each. This node count is aligned with PEMS08 to ensure architectural compatibility, as several baseline models hardcode the number of nodes into their design, making them unsuitable for varying node configurations.

\vspace{-0.5em}
\subsection{Baselines Details}
We conducted a thorough comparison with ten baseline models to ensure a comprehensive evaluation of our proposed approach. These baselines can be broadly categorized into three groups:
\begin{figure}[t]
  \centering
\includegraphics[width=\linewidth]{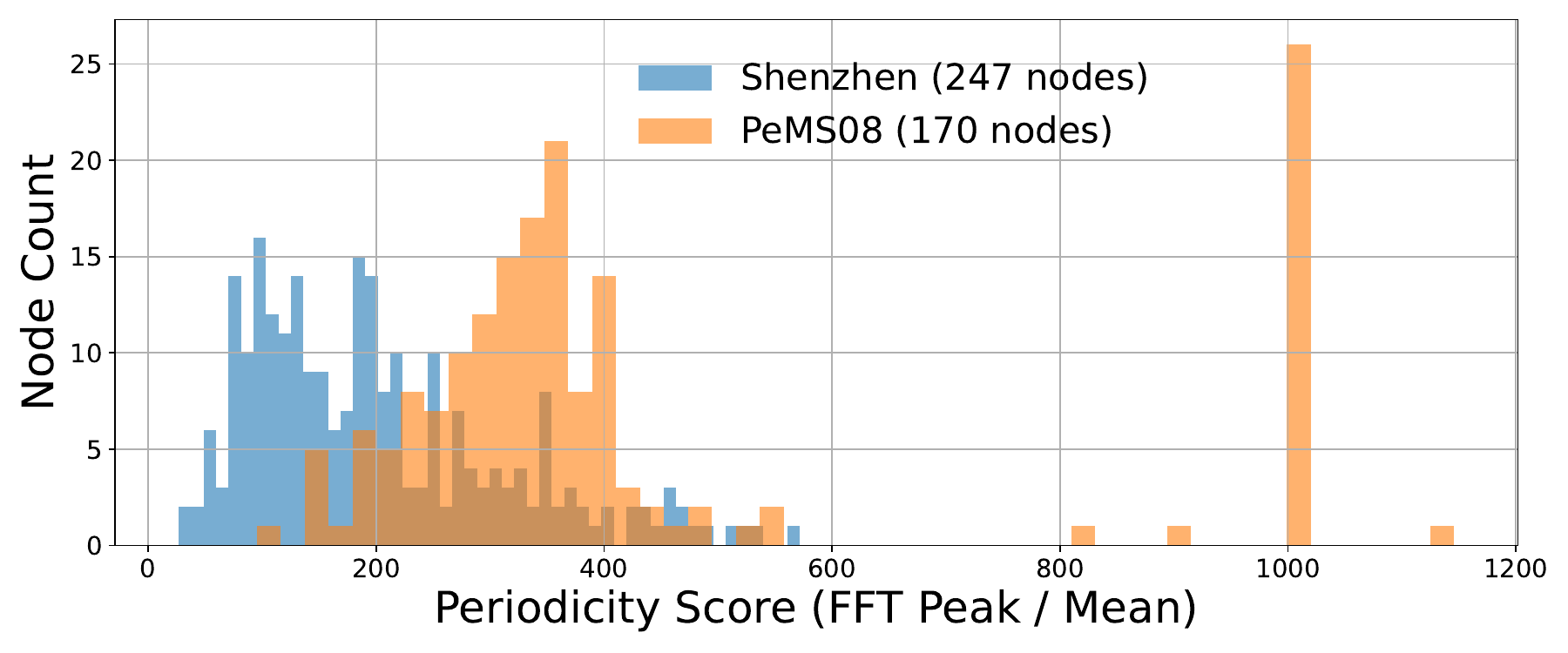}
  \caption{\fontsize{9pt}
  {\baselineskip}\selectfont Periodicity Score Distribution of Charging and Traffic Datasets}
  \vspace{-1.5em}
    \label{fig7}
\end{figure}

\noindent\textbf{1. Small-scale GNN-based Deep Learning Models.}
These models are typically designed for spatiotemporal forecasting and trained in an end-to-end manner on individual datasets. They leverage GNNs to capture spatial dependencies and temporal modeling techniques to understand dynamic patterns over time. This group includes:

\begin{itemize}[leftmargin=10pt]
  \item \textbf{STGCN}~\cite{yu2018spatio}: It combines graph and gated temporal convolutions to capture spatial-temporal patterns in traffic data.
  \item \textbf{DGCRN}~\cite{li2023dynamic}: This framework uses hyper-networks to generate dynamic graphs at each time step and integrates them with static topology to model time-varying traffic correlations.
  \item \textbf{ASTGCN}~\cite{guo2019attention}: It employs spatial and temporal attention mechanisms to focus on salient dependencies across time and space.
  \item \textbf{D$^{2}$STGNN}~\cite{shao2022decoupled}: The framework decouples diffusion and inherent traffic signals and learns dynamic graphs to represent evolving traffic structures more accurately.
  \item \textbf{GWNET}~\cite{wu2019graph}: This model learns an adaptive adjacency matrix and applies dilated convolutions to uncover hidden spatial relations and long-range temporal trends.
  \item \textbf{STGODE}~\cite{fang2021spatial}: It leverages neural ordinary differential equations and semantic adjacency to jointly capture spatial-temporal dynamics.
  \item \textbf{PDG2Seq}~\cite{fan2025pdg2seq}: This method models periodic traffic patterns using a dynamic graph-to-sequence framework that aligns temporal cycles with future trend prediction.
\end{itemize}

\noindent\textbf{2. Generalist Large Language Models (LLMs).}
These are versatile LLMs that have not been specifically trained on spatiotemporal data or transportation tasks. We evaluate them in a zero-shot setting, where no task-specific fine-tuning or supervision is provided:

\begin{itemize}[leftmargin=10pt]
  \item \textbf{Deepseek-v3}~\cite{liu2024deepseek}: It adopts a Mixture-of-Experts architecture to enhance computational efficiency and specialization, and is trained on diverse web-scale data without adaptation to spatiotemporal tasks.
  \item \textbf{GPT-4o}~\cite{hurst2024gpt}: It is a unified multimodal model capable of processing text, vision, and audio inputs, though in our evaluation it is used solely as a zero-shot text-based forecaster without domain-specific tuning.
\end{itemize}

\noindent\textbf{3. Spatiotemporal-Enhanced LLMs.}
This category includes LLM-based architectures that incorporate structured spatiotemporal priors to improve forecasting in urban settings:

\begin{itemize}[leftmargin=10pt]
  \item \textbf{UrbanGPT}~\cite{li2024urbangpt}: It enhances LLMs for urban forecasting by incorporating a temporal encoder that explicitly models periodic patterns and time dependencies in structured urban data.
\end{itemize}

\begin{table*}[t]
    \centering
    \caption{\fontsize{9pt}{\baselineskip}\selectfont \label{tab:case_study}Response Comparison of Different Models on a Specific Traffic Flow Forecasting Instance}
    \vspace{0.3em}
    \setlength{\tabcolsep}{8pt}
    \begin{tabular}{p{17cm}} 
        \toprule
        \textbf{Input Prompt} Given the historical data for traffic flow over 12 time steps at a highway traffic monitoring point, the recorded traffic flow values are [62 67 60 83 81 55 50 56 52 64 41 37]. The recording time of the historical data is 'November 25, 2021, 23:00, Thursday to November 25, 2021, 23:55, Thursday, with data points recorded at 5-minute intervals'. To capture spatial and temporal dependencies, a spatio-temporal convolution model is utilized to encode the historical traffic data as embeddings <ST\_EMB>. Now we want to predict the traffic flow for the next 12 time steps during the time period of 'November 26, 2021, 00:00, Friday to November 26, 2021, 00:55, Friday, with data points recorded at 5-minute intervals'. Please analyze the traffic patterns in this region, taking into account the provided historical data, time encoding, and spatio-temporal embeddings, and generate the predictive tokens for regression in the form "<ST\_PRE>". \\
        \midrule
         \addlinespace[0.5em] 
        \textbf{Personalized Prompt} 'Using traffic flow recorded one hour ago [62, 67, 60, 83, 81, 55, 50, 56, 52, 64, 41, 37], we examine recent short-term variations to identify immediate trends. With this historical context in mind, the model will generate a 12-step forecast beginning at November 26, 2021, 00:00, Friday to November 26, 2021, 00:55, Friday, incorporating 5-minute resolution to synthesize both rapid changes and gradual trends. In order to better understand the early dynamics of traffic flow, we incorporate 12 spatio-temporal embeddings that correspond to future time intervals: <ST\_EMB>. Please reason step-by-step through both temporal patterns and spatial influences. After that, generate the predicted traffic volume for the next 12 time intervals using the token <ST\_PRE>. \\
        \midrule
        \addlinespace[0.5em] 
        \textbf{Ground Truth}  Based on the given historical traffic flow data, time encoding, and spatio-temporal embeddings, the predictive tokens for the traffic flow in this region are:[45, 43, 41, 44, 36, 28, 29, 33, 28, 27, 35, 33] \\
        \midrule
        \addlinespace[0.5em]
        \textbf{Deepseek-v3 Responds:} To analyze the traffic patterns and generate predictive tokens for the next 12 time steps, let\'s break down the problem step-by-step...Based on the above, the predicted traffic flow for the next 12 steps (in 5-minute intervals) is:[58, 62, 59, 78, 79, 54, 49, 54, 50, 62, 40, 36]...\\
        \midrule
        \addlinespace[0.5em]
        \textbf{GPT-4o Responds:} Step 1. Analyze the historical traffic flow...Step 2. Incorporate temporal features... Predict next 12 time steps: [59, 61, 59, 60, 63, 66, 70, 75, 68, 62, 57, 54]... \\
        \midrule
        \addlinespace[0.5em]
        \textbf{UrbanGPT Responds:} Based on the given historical traffic flow data, time encoding, and spatio-temporal embeddings, the predictive tokens for the traffic flow in this region are: [52, 50, 49, 47, 47, 44, 44, 41, 41, 38, 38, 37] \\
        \addlinespace[0.5em]
        \midrule
        \textbf{TransLLM Responds:} Based on the given historical traffic flow data, time encoding, and spatio-temporal embeddings, the predictive tokens for the traffic flow in this region are [49, 47, 38, 42, 33, 32, 33, 29, 29, 28, 29, 31] \\
        \bottomrule
    \end{tabular}
\end{table*}

\vspace{-0.5em}
\subsection{Comparison of PEMS08 and ST-EVCDP Dataset Characteristics}
PEMS08 represents a typical road traffic flow dataset, whereas ST-EVCDP represents a charging demand dataset. These two types of spatio-temporal data exhibit distinct characteristics. Charging demand shows weaker periodicity and greater volatility than traffic flow data. We conduct a Fourier-based periodicity analysis by computing the ratio of the dominant FFT peak to the average signal amplitude (i.e., $\text{FFT Peak} / \text{Mean}$) for each node. As shown in Fig.~\ref{fig7}, the periodicity scores for Shenzhen's 247 charging nodes are significantly lower and more dispersed than those of PeMS08’s 170 traffic sensors. This indicates that while traffic exhibits strong and regular daily or weekly cycles, charging behavior is influenced by more irregular human and behavioral factors—limiting the effectiveness of traditional forecasting models optimized for periodic signals.

\vspace{-0.5em}
\subsection{Case Study}
In this section, we assess the effectiveness of different large models on a representative traffic flow forecasting instance. The detailed input information and the corresponding model responses are summarized in Table~\ref{tab:case_study}. Notably, the 'personalized prompt' is dynamically generated by our Prompt Router for this specific instance. 
Notably, generalist LLMs like Deepseek-v3 and GPT-4o, while capable of parsing the instructions, demonstrate significant limitations. Deepseek-v3 failed to capture the correct temporal patterns, producing a forecast that deviates substantially from the ground truth. GPT-4o adopts a more conservative approach,  producing an overly smooth forecast that misses the inherent volatility. Although UrbanGPT successfully captures the inherent volatility and the overall downward trend, its predictive accuracy is notably lower than that of TransLLM.
